\documentclass[letterpaper]{article}

\usepackage[preprint]{aaai2027}

\usepackage[hyphens]{url}
\usepackage{graphicx}
\usepackage{natbib}
\usepackage{caption}
\usepackage{booktabs}
\usepackage{tabularx}
\usepackage{array}
\usepackage{amsmath,amssymb}

\newcommand{\method}{FinATOM}
\newcommand{\pending}{\textemdash}
\newcommand{\anchor}{\mathrm{A}}
\newcommand{\sharpe}{\operatorname{Sh}}

\newcounter{qacount}
\newcommand{\Q}[1]{%
  \medskip
  \refstepcounter{qacount}%
  \noindent\textbf{Q\theqacount.}\ \emph{#1}\par
  \nopagebreak\smallskip
}
\newcommand{\A}[1]{%
  \noindent\textbf{A.}\ #1\par
}

\title{\method: Financial Numerical Prediction and Allocation as Token Generation}

\author{
Xu Ouyang, Moontae Lee
}

\affiliations{
University of Illinois Chicago\\
xouya@uic.edu, moontae@uic.edu
}

\begin{document}
\maketitle

\begin{abstract}
Financial prediction typically relies on task-specific regression, ranking, or policy heads, separating the language model from the numerical object ultimately evaluated. We investigate whether a causal language model can instead represent forecasts and decisions directly through constrained token generation. FinATOM introduces a unified, head-free interface for three-step stock-return forecasting and dynamic five-ETF allocation. The forecasting model autoregressively emits volatility-standardized return tokens and is trained with ordinal and ranking supervision followed by a one-epoch token-level policy stage. The allocation model generates normalized long-only weights; supervised fine-tuning imitates a causal mean--variance anchor, and DAPO-augmented GRPO optimizes realized 21-day Sharpe subject to anchor consistency. In 2023--2025 ETF tests, the allocation policy improves pooled gross Sharpe from 1.428 to 1.529 and net Sharpe under a 5-bp transaction-cost model from 1.394 to 1.494. The multimodal allocation input attains the highest three-period mean Sharpe of 1.540, with its clearest advantage in 2025. On FinTexTS, the SFT and policy strategies achieve 73.52\%/2.68 and 73.72\%/2.69 cumulative-return/Sharpe, respectively. These results support the feasibility of direct language-model token generation for financial numerical prediction and decision-making, while motivating broader tests across assets, regimes, and random seeds.
\end{abstract}

\begin{figure}[t]
\centering
\includegraphics[width=0.90\columnwidth]{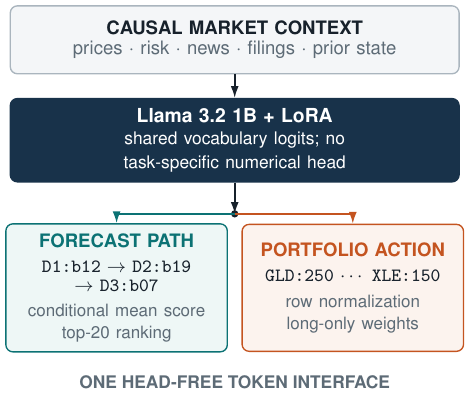}
\caption{\method{}'s head-free interface: one causal language model emits constrained forecast or allocation tokens without a task-specific numerical head.}
\label{fig:overview}
\end{figure}

\section{Introduction}

Financial systems usually combine heterogeneous evidence to produce numerical
decisions. Prices describe market dynamics, while news and regulatory
filings contain information about monetary policy, industry conditions,
firm fundamentals, and events. The required outputs also differ:
forecasting produces scalar or multi-step returns, whereas portfolio
construction produces a jointly constrained vector. Most systems encode
the context and attach a task-specific regression, ranking, or policy
head \cite{nie2023patchtst,li2024master,jin2024timellm}. Although
effective, this fragments the output contract: each task needs its own
parameterization, loss, decoder, and validity mechanism, and the
language model does not directly represent the final financial object.

\newpage

We study whether the language model itself can serve as the numerical
decision interface. Direct generation offers several attractive
properties: the output is auditable, mechanically parseable, and
compatible with both supervised fine-tuning and token-level policy
optimization. Nevertheless, numerical token generation is not a
trivial replacement for a regression head. Ordinary number strings
have irregular tokenization and do not preserve numerical order;
unconstrained decoding can produce malformed forecasts or portfolio
weights that violate long-only and budget constraints; supervised
targets can inadvertently incorporate future information; and token
cross-entropy alone need not optimize the return--risk criterion used
in deployment. A credible token-native design therefore requires an
ordered numerical vocabulary, an explicit output grammar, a
continuous decoding rule, causal supervision, and an objective aligned
with the downstream financial decision.

\method{} provides this design without adding a task-specific
numerical head to the standard language-model vocabulary projection.
As summarized in Figure~\ref{fig:overview}, the same causal Llama 3.2
1B backbone supports two complementary tasks. For stock forecasting,
the prompt combines a 64-session price-derived table with market
context, SEC-filing summaries, and FinTexTS news organized at macro,
sector, related-company, and target-company levels
\cite{lee2026fintexts}. The model autoregressively emits ordered tokens
for the consecutive D1, D2, and D3 returns. Conditional expectations
under the token distributions recover continuous return scores for
cross-sectional ranking while preserving the discrete generation
trace. For portfolio allocation, the prompt combines a visible
20-day return window, risk and correlation statistics, macroeconomic
news, and the previous portfolio state. The model directly emits five
long-only ETF weights, whose normalized values constitute the
deployed action.

The allocation task further separates causal imitation from
outcome-based improvement. Supervised fine-tuning learns a
mean--variance anchor computed only from information observable at
the decision date, followed by shrinkage and deterministic
quantization. A subsequent DAPO-augmented GRPO stage
\cite{shao2024deepseekmath,yu2025dapo} samples legal allocation
sequences and rewards their realized 21-day Sharpe while penalizing
excessive deviation from the causal anchor. Future returns enter only
this policy reward: they are absent from both the prompt and the
supervised target. The language model therefore first learns a stable,
fully specified allocation grammar and is then optimized against the
financial objective directly, without introducing a separate policy,
value, or reward head.

\method{} complements several related research directions. Prior work
serializes numerical sequences for language-model forecasting
\cite{gruver2023llmtime}, develops tokenized time-series models
\cite{tao2025tokencast,shi2025kronos}, generates quantized structured
outputs \cite{chen2022pix2seq,jiang2026detect}, and applies language
models to financial prediction and decision-making
\cite{koa2024selfreflective}. The distinguishing question studied here
is whether a single constrained, head-free output interface can
support both forecasting and allocation despite their different
numerical semantics, validity constraints, and training signals.
Forecasting evaluates whether the interface can encode an ordered
future path, whereas allocation evaluates whether it can generate a
jointly constrained action and improve that action through
task-aligned policy optimization.

Chronological ETF experiments over 2023--2025 show that the policy
stage raises pooled gross Sharpe from 1.428 to 1.529 and net Sharpe
under a 5-bp transaction-cost model from 1.394 to 1.494. The
input-modality ablation yields the highest three-period mean Sharpe
with news and time series together, although the multimodal advantage
is concentrated in 2025 rather than uniform across years. On FinTexTS,
SFT attains 73.52\% cumulative return and 2.68 Sharpe, and the one-epoch
policy stage reaches 73.72\% and 2.69. The FinATOM SFT-to-policy rows
share one protocol, whereas comparison with the published FinTexTS
system remains contextual. Together, the experiments suggest that
direct token generation can represent both financial forecasts and
constrained decisions, while exposing important questions in
calibration and nonstationarity.

Our contributions are threefold:
\begin{itemize}
    \item We introduce a shared constrained token interface for
    autoregressive three-step return forecasting and normalized
    five-ETF allocation, without an additional task-specific numerical
    head.

    \item We develop a causal allocation pipeline that combines a
    fully specified mean--variance teacher, ordinal token supervision,
    and DAPO-augmented GRPO aligned with realized portfolio Sharpe.

    \item We provide chronological allocation evidence across three
    test periods, a controlled input-modality ablation, and an
    exploratory FinTexTS SFT-to-policy comparison, while separating
    within-protocol evidence from contextual comparison with the
    published baseline.
\end{itemize}

\section{Related Work}
\paragraph{Language models and time series.}
LLMTime serializes numerical sequences directly \cite{gruver2023llmtime}; One Fits All and Time-LLM adapt pretrained language models to time-series inputs \cite{zhou2023one,jin2024timellm}. Specialized architectures such as PatchTST and iTransformer remain strong alternatives \cite{nie2023patchtst,liu2024itransformer}, and recent work questions when language-model priors improve forecasting \cite{tan2024useful}. OpenTSLM integrates time series with a language-model interface through multimodal adaptation \cite{langer2026opentslm}. TokenCast and Kronos further illustrate tokenized or pretrained sequence modeling for time series \cite{tao2025tokencast,shi2025kronos}. Our focus is the output side: the forecast or action is generated through the LM vocabulary rather than a separate numerical head.

\paragraph{Financial text and prediction.}
FNSPID and Time-MMD provide text--time-series resources \cite{dong2024fnspid,liu2024timemmd}. FinTexTS pairs stock histories with macro, sector, related-company, and target-company text \cite{lee2026fintexts}. MASTER models temporal and cross-sectional stock structure with a specialized network \cite{li2024master}; language-model studies also find predictive information in headlines, filings, and self-reflective decision traces \cite{lopezlira2023chatgpt,kim2024financial,koa2024selfreflective}. We use FinTexTS only as an exploratory stock-ranking setting and keep it separate from the controlled ETF experiment.

\paragraph{Structured generation and policy optimization.}
Pix2Seq and next-point prediction show that quantized numerical structures can be learned as token sequences \cite{chen2022pix2seq,jiang2026detect}. Mean--variance optimization supplies a causal allocation teacher \cite{markowitz1952portfolio}, while equal weighting is a difficult out-of-sample reference \cite{demiguel2009optimal}. GRPO estimates critic-free group-relative advantages \cite{shao2024deepseekmath}. DAPO extends this family with decoupled Clip-Higher, dynamic sampling, token-level policy gradients, and overlong reward shaping \cite{yu2025dapo}; a financial study likewise combines improved GRPO with DAPO ideas \cite{zha2025stockdapo}. Our allocation stage adopts the applicable asymmetric clipping and token-level update, but not the complete DAPO recipe.

\begin{figure*}[t!]
\centering
\includegraphics[width=\textwidth]{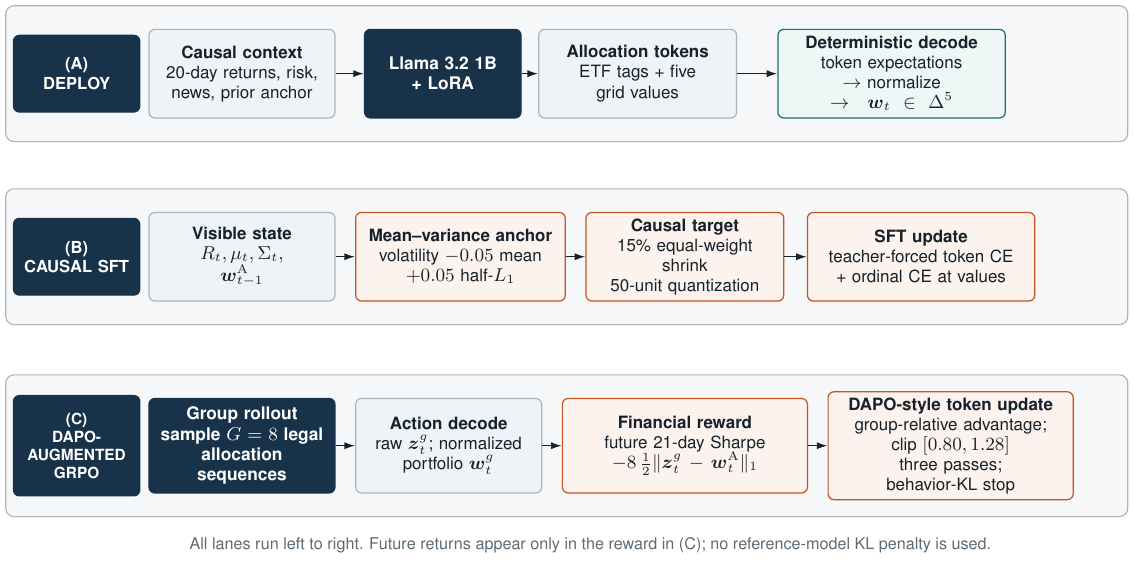}
\caption{Allocation architecture. Each lane proceeds left to right. The SFT target is constructed only from visible information, whereas forward 21-day returns enter only the policy reward. The policy update uses group-relative advantages and DAPO-style asymmetric token clipping, with no reference-model KL penalty.}
\label{fig:portfolio_arch}
\end{figure*}

\section{Method}
\begin{figure*}[t!]
\centering
\includegraphics[width=\textwidth]{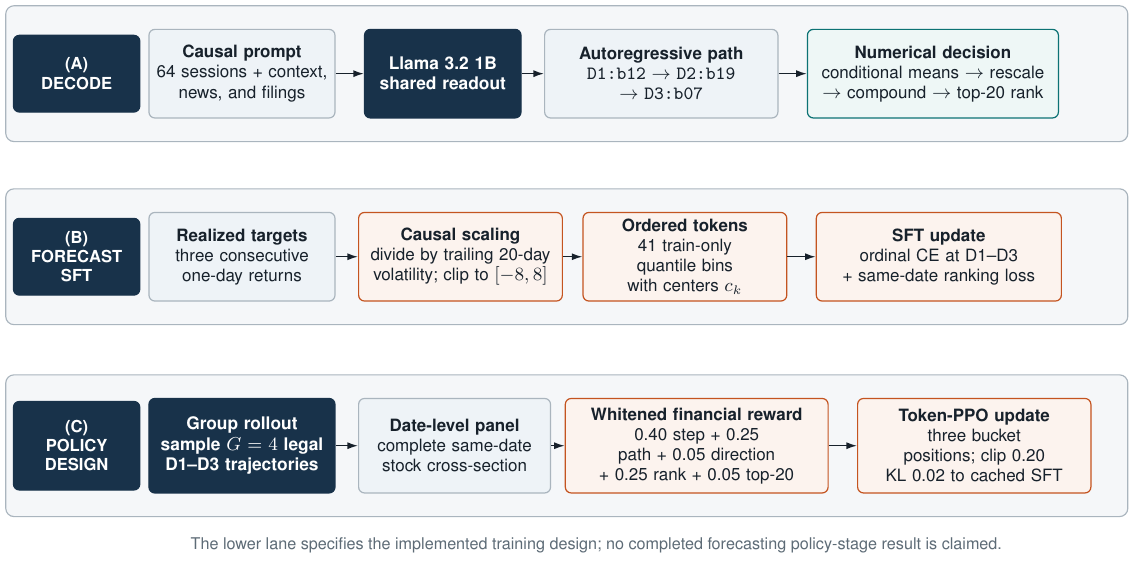}
\caption{Forecasting architecture. D1, D2, and D3 are generated autoregressively; conditional token means produce the continuous ranking score. The lower lane shows the evaluated one-epoch policy stage.}
\label{fig:forecast_arch}
\end{figure*}

\paragraph{Model overview.}
Figures~\ref{fig:portfolio_arch} and~\ref{fig:forecast_arch} show the allocation and forecasting instantiations of the shared token-native design. Both instantiate the same LoRA-adapted Llama 3.2 1B architecture and tied vocabulary projection, but use task-specific output grammars and training objectives. Target construction and reward computation are external training operations rather than additional neural heads; inference retains only the causal prompt, constrained token generation, and deterministic numerical decoding.

\subsection{Shared Token-Native Interface}
For causal context $x_t$ and structured answer $y_t=(y_{t,1},\ldots,y_{t,L})$, the backbone models
\begin{equation}
 p_\theta(y_t\mid x_t)=\prod_{\ell=1}^{L}p_\theta(y_{t,\ell}\mid x_t,y_{t,<\ell}).
 \label{eq:lm}
\end{equation}
Both tasks adapt Llama 3.2 1B \cite{dubey2024llama} with LoRA \cite{hu2022lora}. Valid numerical values are learned vocabulary items scored by the tied LM readout. No regression, policy, value, or reward head is appended. Table~\ref{tab:task_summary} contrasts the two contracts.

\begin{center}
\small
\setlength{\tabcolsep}{3pt}
\renewcommand{\arraystretch}{1.10}
\begin{tabularx}{\columnwidth}{
    >{\bfseries}p{0.18\columnwidth}
    X
    X
    X
}
\toprule
Task & Token output & SFT objective & Decode \\
\midrule
Allocation
& ETF tags + five grid values
& Token CE + ordinal CE at value slots
& Expectation decode; normalize row \\
Forecast
& D1--D3 tokens from 41 bins
& Ordinal CE + same-date rank loss
& Autoregressive prefix; expectation per slot \\
\bottomrule
\end{tabularx}

\captionof{table}{Two numerical contracts implemented through the same causal token interface.}
\label{tab:task_summary}
\end{center}

\subsection{Portfolio: Causal Mean--Variance Target}
\paragraph{Allocation architecture.}
Figure~\ref{fig:portfolio_arch} separates deployment, causal SFT, and policy optimization. At deployment, the model maps the serialized 20-day market context and previous anchor to ETF tags and five grid-valued tokens, whose expectations are normalized into a long-only portfolio. The SFT lane constructs the mean--variance anchor exclusively from the visible state, whereas the policy lane samples legal allocations and evaluates them using forward 21-day Sharpe together with the anchor penalty. Both objectives update the same language-model parameters; no separate allocation, value, or reward network is introduced.

The universe is ordered as GLD, SPY, TLT, UUP, and XLE. Let $R_t\in\mathbb{R}^{20\times5}$ be the strictly visible return window ending at decision date $t$, with sample mean $\mu_t$ and covariance $\Sigma_t$. The continuous teacher solves
\begin{align}
\widetilde{\boldsymbol w}_t=\arg\min_{\boldsymbol w\in\mathcal W}\;&
\sqrt{252\,\boldsymbol w^\top\Sigma_t\boldsymbol w}
-0.05(252\mu_t)^\top\boldsymbol w \notag\\
&+0.05\,\frac12\lVert\boldsymbol w-\boldsymbol w^{\anchor}_{t-1}\rVert_1,
\label{eq:anchor}
\end{align}
where $\mathcal W=\{\boldsymbol w\ge0,\ \boldsymbol 1^\top\boldsymbol w=1,\ w_i\le0.5\}$. The first date omits the turnover term. Thus the target is causal but path dependent: the previous quantized anchor is both prompt state and optimization state.

The solution is shrunk and discretized as
\begin{equation}
\overline{\boldsymbol w}_t=0.85\widetilde{\boldsymbol w}_t+0.15\boldsymbol 1/5,
\quad
\boldsymbol a_t^{\anchor}=Q_{50,1000}(1000\overline{\boldsymbol w}_t),
\label{eq:quant}
\end{equation}
where $Q_{50,1000}$ uses largest-remainder allocation on a 50-unit grid, enforces a 1000-unit budget, and assigns at least one grid step to every asset. The normalized state $\boldsymbol w_t^{\anchor}=\boldsymbol a_t^{\anchor}/1000$ is passed to the next date. This is a designed causal teacher, not future-return ground truth.

\subsection{Portfolio SFT and DAPO-Augmented GRPO}
The prompt contains the 20-day return table, annualized risk and correlations, calendar fields, daily macro-financial news, asset order, and $\boldsymbol w_{t-1}^{\anchor}$. A target answer has the form
\begin{center}
\small\texttt{<GLD><250><SPY><200>...<XLE><150>}.
\end{center}
Every boundary, ETF tag, and value receives sequence cross-entropy. At the five value positions, an additional Gaussian ordinal target over $\{0,50,\ldots,1000\}$ is used:
\begin{equation}
\begin{aligned}
 q_{jk}&\propto\exp\!\left[-\frac{(k-k_j^*)^2}{2(0.8)^2}\right],\\
\mathcal L_{\mathrm{SFT}}&=\mathcal L_{\mathrm{tok}}+\sum_{j=1}^{5}\mathrm{CE}(q_j,p_j).
\end{aligned}
\label{eq:alloc_sft}
\end{equation}
At inference, $\widehat a_j=\sum_kp_{jk}v_k$ and the row is normalized to sum to one; a degenerate row falls back to equal weight.

The policy stage samples $G=8$ legal allocation sequences. For raw token action $\boldsymbol z_t^g=\boldsymbol a_t^g/1000$, deployed weights are $\boldsymbol w_t^g=\boldsymbol z_t^g/(\boldsymbol1^\top\boldsymbol z_t^g)$. Using only returns $t+1{:}t+21$, the reward is
\begin{equation}
 R_t^g=\sharpe(\boldsymbol w_t^g;\,t+1{:}t+21)
 -8\,\frac12\lVert\boldsymbol z_t^g-\boldsymbol w_t^{\anchor}\rVert_1,
\label{eq:alloc_reward}
\end{equation}
with invalid parses assigned $-5$. Group advantages are
\begin{equation}
 A_t^g=\frac{R_t^g-\overline R_t}{\max(s_t,0.65)},
 \label{eq:group_adv}
\end{equation}
and are set to zero for effectively constant-reward groups. For active answer token $\ell$, let $\rho_{t\ell}^g=\pi_\theta/\pi_{\mathrm{old}}$. The token-level clipped objective is
\begin{equation}
\mathcal L_{\mathrm{pol}}=-\frac{1}{M}\!\sum_{g,\ell}\!m_{g\ell}
\min\!\left(\rho_{t\ell}^g A_t^g,
\operatorname{clip}(\rho_{t\ell}^g,0.80,1.28)A_t^g\right).
\label{eq:policy_loss}
\end{equation}
Training uses three passes and behavior-KL early stopping at 0.03, but the reference-KL coefficient is zero. This is best described as \emph{DAPO-augmented GRPO}: it combines group-relative advantages with DAPO's decoupled Clip-Higher and token-level update, but does not implement dynamic batch replenishment or overlong reward shaping \cite{yu2025dapo}. Neither the loss nor any KL term compares predictions with future target weights.

\subsection{Forecast Targets, SFT, and Autoregressive Decode}
\paragraph{Forecasting architecture.}
Figure~\ref{fig:forecast_arch} organizes the forecasting model into autoregressive decoding, supervised target construction, and policy refinement. The causal prompt is processed by the same backbone architecture, which emits D1, D2, and D3 bucket tokens sequentially so that each generated prefix conditions the next horizon. Conditional means over the ordered bucket distributions are rescaled by trailing volatility and compounded into the continuous score used for top-20 ranking. SFT learns train-only, volatility-standardized bins, while the one-epoch policy stage samples complete paths and updates the same three bucket positions using the five-component financial reward.

The forecasting prompt contains 64 sessions of return, gap, and range features; risk, return, and drawdown summaries; market and sector context; four levels of paired news; and filing summaries. For $h\in\{1,2,3\}$, the target is the consecutive one-day return
\begin{equation}
\begin{aligned}
 r_{t,h}&=P_{t+h}/P_{t+h-1}-1,\\
 z_{t,h}&=\operatorname{clip}(r_{t,h}/\sigma_{t,20},-8,8),
\end{aligned}
\label{eq:forecast_target}
\end{equation}
where $\sigma_{t,20}$ is trailing 20-day volatility known at $t$. Forty-one quantile bins are fit on training targets pooled across training tickers and horizons; each bin has center $c_k=\mathbb E[z\mid b=k]$. The fixed grammar is
\begin{center}
\small\texttt{<|forecast\_start|><D1><b12><D2><b19>}\\[-1mm]
\small\texttt{<D3><b07><|forecast\_end|>}.
\end{center}
SFT computes Gaussian ordinal CE only at the three bucket slots and adds a same-date, per-horizon pairwise ranking loss with coefficient 0.5. The scaffold is fixed rather than learned as ordinary answer text.

Decoding is genuinely autoregressive. The D1 argmax bucket is appended before evaluating D2, and D2 is appended before D3. Numerical values remain conditional expectations, $\widehat z_{t,h}=\sum_kp_{t,h,k}c_k$, then are rescaled by $\sigma_{t,20}$ and compounded into a three-day ranking score. Thus the discrete prefix and continuous score serve complementary roles.

\subsection{Forecast Policy Optimization}
The evaluated one-epoch policy stage samples $G=4$ legal autoregressive trajectories and forms complete decision-date stock cross-sections. Five reward components are whitened within each group and combined as
\begin{equation}
\begin{aligned}
R={}&0.40R_{\mathrm{step}}+0.25R_{\mathrm{path}}+0.05R_{\mathrm{dir}}\\
&+0.25R_{\mathrm{rank}}+0.05R_{\mathrm{top20}}.
\end{aligned}
\label{eq:forecast_reward}
\end{equation}
A clipped token-PPO update is applied only at the three sampled bucket positions. KL coefficient 0.02 compares current action probabilities with cached source-SFT probabilities; no second reference network or reward head is instantiated. This mechanism is distinct from the allocation stage's zero-reference-KL DAPO-augmented GRPO. The resulting checkpoint is evaluated against its SFT initialization under the same stock-ranking and backtesting protocol.

\begin{table*}[t]
\centering
\small
\setlength{\tabcolsep}{7pt}
\renewcommand{\arraystretch}{1.05}
\begin{tabular}{@{}llrrrr@{}}
\toprule
Year & Method & Ann. return & Ann. vol. & Sharpe & Max drawdown \\
\midrule
2023 & SFT & 7.18\% & 5.03\% & 1.404 & -2.81\% \\
     & Policy & 7.11\% & 4.98\% & \textbf{1.404} & -2.78\% \\
     & Causal target & 7.92\% & 5.03\% & 1.542 & -2.83\% \\
     & Equal weight & 6.55\% & 7.47\% & 0.886 & -5.28\% \\
\midrule
2024 & SFT & 7.35\% & 5.00\% & 1.445 & -4.72\% \\
     & Policy & 7.88\% & 4.93\% & \textbf{1.564} & -4.72\% \\
     & Causal target & 7.06\% & 4.93\% & 1.407 & -4.69\% \\
     & Equal weight & 9.59\% & 6.87\% & 1.367 & -4.88\% \\
\midrule
2025 & SFT & 10.46\% & 6.91\% & 1.474 & -6.28\% \\
     & Policy & 11.76\% & 6.87\% & \textbf{1.653} & -6.04\% \\
     & Causal target & 11.58\% & 6.61\% & 1.691 & -5.21\% \\
     & Equal weight & 15.72\% & 10.03\% & 1.506 & -7.99\% \\
\bottomrule
\end{tabular}
\caption{Out-of-sample five-ETF results. ``Policy'' denotes DAPO-augmented GRPO initialized from SFT. Return and volatility are annualized; the causal target is a numerical reference rather than a learned policy. The 2025 test ends October 31.}
\label{tab:portfolio_results}
\end{table*}

\section{Experimental Setup}
\paragraph{ETF allocation dataset.}
We construct a daily multimodal dataset for five liquid ETFs: GLD, SPY, TLT, UUP, and XLE, from January 2018 to October 2025. The underlying market data contain daily OHLCV observations. For each decision date $t$, the LLM receives a chronological text table covering the latest 20 trading days, with the relative day, calendar date, and each ETF's close-to-close return expressed as integer basis points. The prompt also includes annualized volatility, the $5\times5$ return-correlation matrix computed from the same window, and the previous portfolio allocation. We collect macroeconomic news through NewsAPI \cite{newsapi2026} from major publishers, including \emph{Bloomberg, Reuters, CNBC} and other widely followed outlets. Articles concerning monetary policy, inflation, growth, employment, trade, energy, and geopolitical risk are grouped by date and summarized by DeepSeek-V4 \cite{deepseek2026v4} into an approximately 100-word daily news block.

\paragraph{FinTexTS forecasting dataset.}
For stock forecasting, we use FinTexTS \cite{lee2026fintexts}, which covers 100 large U.S.\ companies from 2019 to 2023 and pairs their price histories with text derived from approximately one million news articles and SEC filings. Its textual information is organized into macro-, sector-, related-company-, and target-company-level blocks. For each stock and decision date, the LLM receives the latest 64 trading sessions as a text table containing close-to-close returns, opening gaps, and intraday high--low ranges in integer basis points. The prompt additionally reports trailing volatility, multi-horizon returns, drawdown from the 52-week high, and same-day market and sector context. The four news levels and five SEC-filing categories are retained as separately labeled text blocks, after which the model autoregressively emits three tokens representing the consecutive returns on days D1, D2, and D3.

\paragraph{Portfolio allocation.}
Three expanding chronological experiments train on 2018--2022 and test 2023 (249 days), train 2019--2023 and test 2024 (251 days), and train 2020--2024 and test January--October 2025 (209 days). SFT runs five epochs with batch size 1, gradient accumulation 4, learning rate $2\times10^{-4}$, bfloat16, LoRA rank 64, scale 128, dropout 0.05, ordinal width 0.8, and ordinal coefficient 1. The policy stage runs one epoch with learning rate $3\times10^{-5}$, group size 8, and temperature 1. The modality ablation reruns this same SFT-to-policy pipeline while changing only the included market modalities; dates, asset order, and the previous anchor remain available in every arm. The final 20 training dates are excluded because they lack a fully contained future reward window. Metrics use daily returns, zero risk-free rate, and 252-day annualization. Pooled evaluation concatenates all 709 test days; net Sharpe charges 5 basis points per unit of one-way turnover.

\paragraph{Stock forecasting.}
The universe contains 100 FinTexTS equities. Context and bucket construction use training observations before 2022, 2022 is validation, and the reported test contains 239 trading days through December 19, 2023. Stocks are ranked by decoded three-day return; the top 20 are equally weighted and rebalanced every three sessions with 5-bp one-way costs. We evaluate both SFT and its one-epoch policy refinement. The policy stage uses group size 4, a policy learning-rate decay from $10^{-6}$ to $10^{-7}$, and KL coefficient 0.02 to cached SFT action probabilities.

\paragraph{Chronology and leakage controls.}
Allocation prompts contain observations stamped no later than $t$. The causal target is recomputed sequentially inside each experiment, so the previous-anchor state cannot cross from test into training. Reward windows begin at $t+1$ and are separate from prompt fields; terminal training dates without 21 future sessions are dropped. Forecast volatility, bin boundaries, and bin centers are estimated from training data only, and ranking losses compare stocks sharing the same decision date. The base-model pretraining cutoff is discussed below.

\paragraph{Evaluation and comparison scope.}
For portfolio $\boldsymbol w_t$, next-session gross return is $\boldsymbol w_t^\top\boldsymbol r_{t+1}$. One-way turnover is $\tfrac12\lVert\boldsymbol w_t-\boldsymbol w_{t-1}\rVert_1$; net return subtracts $0.0005$ times turnover. Annual return is daily mean times 252, volatility is daily standard deviation times $\sqrt{252}$, and Sharpe is their ratio. Deterministic expectation decoding is used throughout. The causal target and equal weight are numerical references, while the allocation SFT-to-policy comparison holds the backbone, grammar, context, and decoder fixed. In the stock experiment, FinATOM SFT and Policy also share a common split, decoder, and backtest; only their comparison with the published FinTexTS row is contextual rather than controlled.

\section{Results}
\subsection{Portfolio Allocation}
\begin{figure}[t]
\centering
\includegraphics[width=\columnwidth]{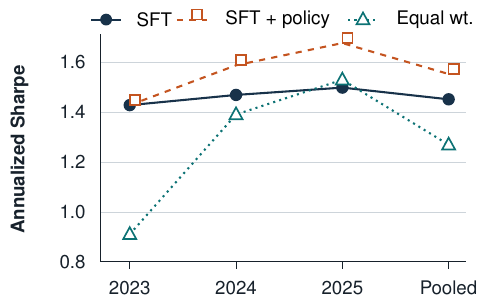}
\caption{Sharpe by test period and after pooling all 709 days. The policy curve is DAPO-augmented GRPO initialized from SFT.}
\label{fig:portfolio_sharpe}
\end{figure}

Table~\ref{tab:portfolio_results} shows that the policy stage is unchanged from SFT at three-decimal Sharpe in 2023, but improves both sides of the ratio in 2024 and 2025. Annual return rises by 0.52 and 1.31 percentage points, respectively, while volatility falls by 0.07 and 0.04 points. The 2024 policy also exceeds the causal teacher's Sharpe (1.564 versus 1.407), indicating that policy optimization is not merely reproducing the SFT target; the reversal in 2023 and 2025 shows that the gain remains regime dependent. Equal weighting earns more raw return in 2024--2025, but at substantially higher volatility and larger drawdown. Figure~\ref{fig:portfolio_sharpe} visualizes this regime-dependent pattern: the policy matches SFT in 2023, separates positively in 2024 and 2025, and achieves the strongest pooled Sharpe among the learned policies.

The pooled comparison in Table~\ref{tab:pooled} avoids averaging annual ratios. Policy optimization raises gross Sharpe by 0.101 and net Sharpe by 0.100, while average one-way turnover increases only from 1.481\% to 1.537\% (5.6 basis points per day). Its gross Sharpe of 1.529 is within 0.011 of the causal target's 1.540. Together, these results suggest that the policy stage mainly improves the return--risk tradeoff rather than exploiting extra trading intensity.

\begin{center}
{\small
\setlength{\tabcolsep}{4pt}
\begin{tabular}{lrrrrr}
\toprule
Policy & 2023 & 2024 & 2025 & Pooled & Net 5 bp \\
\midrule
SFT & 1.404 & 1.445 & 1.474 & 1.428 & 1.394 \\
Policy & 1.404 & \textbf{1.564} & \textbf{1.653} & \textbf{1.529} & \textbf{1.494} \\
Causal target & 1.542 & 1.407 & 1.691 & 1.540 & \pending \\
Equal weight & 0.886 & 1.367 & 1.506 & 1.245 & \pending \\
\bottomrule
\end{tabular}}
\captionof{table}{Portfolio Sharpe across test periods. Bold indicates improvement of the learned policy over its SFT initialization. Net values are reported where the same turnover-cost calculation is available.}
\label{tab:pooled}
\end{center}

\subsection{Input-Modality Ablation}
Table~\ref{tab:modality_ablation} isolates the contribution of news and time-series context using the same SFT-to-policy training pipeline, while retaining dates, asset order, and the previous causal anchor in every arm. The combined input obtains the highest simple mean Sharpe, 1.540, exceeding time series only by 0.061 and news only by 0.104. The annual pattern is not uniform: time series only leads the combined model by just 0.004 in 2023 and 0.001 in 2024, whereas news plus time series leads by 0.190 over time series only and by 0.180 over news only in 2025. Thus, recent returns and risk statistics provide the more stable standalone signal, while macroeconomic news appears complementary when the 2025 regime makes it informative. Because the table contains only three single-run periods, the result supports regime-dependent complementarity rather than a statistically general claim that adding news always helps.

\begin{center}
{\small
\setlength{\tabcolsep}{4pt}
\begin{tabular}{lrrrr}
\toprule
Input modality & 2023 & 2024 & 2025 & Mean Sharpe \\
\midrule
News only
    & 1.377
    & 1.459
    & 1.473
    & 1.436 \\
Time series only
    & \textbf{1.408}
    & \textbf{1.565}
    & 1.463
    & 1.479 \\
News + time series
    & 1.404
    & 1.564
    & \textbf{1.653}
    & \textbf{1.540} \\
\bottomrule
\end{tabular}}
\captionof{table}{Input-modality ablation for the full SFT-to-policy pipeline, measured by annual Sharpe ratio. Dates and the previous causal anchor remain present in every arm; the final column is the simple mean across the three test periods. Bold values indicate the best result in each column.}
\label{tab:modality_ablation}
\end{center}

\subsection{Exploratory Stock Result}

\begin{center}
{\small
\setlength{\tabcolsep}{8pt}
\begin{tabular}{lrr}
\toprule
System & Cumulative return & Sharpe \\
\midrule
FinTexTS & 53.98\% & 2.67 \\
\method{} SFT & 73.52\% & 2.68 \\
\method{} Policy & \textbf{73.72\%} & \textbf{2.69} \\
\bottomrule
\end{tabular}}
\captionof{table}{Exploratory 2023 stock-strategy results. FinATOM SFT and Policy use the same protocol; comparison with the published FinTexTS system is contextual because its protocol differs.}
\label{tab:fintexts_backtest}
\end{center}

Within the common FinATOM protocol, the one-epoch policy stage raises cumulative return from 73.52\% to 73.72\% and Sharpe from 2.68 to 2.69. The small increment indicates that most of the strategy performance is already learned during SFT; policy optimization provides a modest refinement rather than a wholesale change in the ranking signal. Relative to the published FinTexTS row, the policy strategy is higher by 19.74 percentage points in cumulative return (36.6\% relative) but only 0.02 in Sharpe. This divergence between terminal wealth and risk-adjusted efficiency suggests that the large return gap may reflect compounding, exposure, timing, or other protocol differences rather than uniformly superior forecasts. Because the published baseline does not share the same split and the table lacks matched volatility, drawdown, and turnover, it does not establish cross-system dominance. The supported conclusions are narrower: the head-free token score sustains a high-Sharpe top-20 strategy, and the evaluated policy stage improves it slightly without destabilizing performance.

\section{Discussion and Limitations}
The allocation results support a more specific interpretation than a uniform benefit from reinforcement learning. DAPO-augmented GRPO is effectively neutral in 2023 and improves Sharpe in 2024 and 2025; the pooled net gain survives the 5-bp cost model while turnover rises only slightly. The policy stage therefore appears to refine the return--risk tradeoff rather than obtain performance simply by trading more. 

The stock experiment separates two claims. The within-protocol SFT-to-policy comparison improves both reported metrics slightly, suggesting that the supervised token representation carries most of the predictive signal and the policy objective provides only incremental task alignment. By contrast, the larger cumulative-return difference relative to FinTexTS is not accompanied by a comparably large Sharpe difference.

The study remains limited to five ETFs, three approximately one-year allocation tests, a partial 2025 period, one training seed per setting, and stylized transaction costs; accordingly, the results should be interpreted as evidence of feasibility rather than universal robustness. Meta reports a December 2023 pretraining-data cutoff for Llama 3.2 \cite{meta2024llama32card}, and its model card does not identify targeted pretraining on our daily ETF data, macro-news summaries, or time-series-to-allocation task. In particular, the market observations and news used in the 2024 and 2025 allocation tests occur after the disclosed cutoff, reducing the likelihood that improvements in these periods arise from direct memorization of the evaluated events. At the same time, the base model may still contribute broad financial and economic knowledge acquired during pretraining, while the 2023 result warrants more cautious interpretation because it overlaps the cutoff year. Future work should examine whether these findings persist across broader asset universes, longer horizons, additional market regimes, and repeated training seeds.

\section{Conclusion}
\method{} demonstrates that a causal language model can emit both multi-step return forecasts and constrained portfolio allocations through one head-free token interface. For allocation, a strictly causal mean--variance teacher supplies supervised targets and DAPO-augmented GRPO improves pooled gross and net Sharpe with little additional turnover. The modality ablation shows that time-series context provides the stable foundation, while news contributes most strongly in the 2025 regime. For forecasting, the same vocabulary-based interface supports autoregressive D1--D3 scores, and the evaluated one-epoch policy stage modestly improves the SFT strategy from 73.52\% return and 2.68 Sharpe to 73.72\% and 2.69. These findings establish feasibility rather than universal superiority. Broader asset universes, longer post-cutoff evaluation periods, repeated seeds, matched baselines, and uncertainty-aware analyses will further strengthen the evidence for the generality and robustness of direct numerical token generation in financial modeling.

\bibliography{references}

\clearpage
\setcounter{qacount}{0}
\setcounter{section}{0}

\begin{center}
{\LARGE\bfseries Supplementary Material}
\end{center}
\vspace{0.5em}

\section{Scope, Motivation, and Novelty}

\Q{Why use a 1B-parameter language model for a five-asset allocation problem that a tiny network, or the mean--variance optimizer itself, can solve?}
\A{The research question is not parameter efficiency on this specific problem but whether a single causal LM can act as the numerical decision interface for heterogeneous financial outputs---an ordered multi-step forecast and a jointly constrained simplex action---while ingesting text and numbers in one prompt. The five-ETF setting is a deliberately small, controlled testbed in which a fully specified causal teacher exists, so imitation and outcome-based improvement can be separated cleanly. The same interface then scales to the 100-stock forecasting task, whose four-level news and filing text cannot be consumed by compact numerical models at all.}

\Q{The learned policy's pooled Sharpe (1.529) trails the causal target (1.540). Why not simply deploy the anchor?}
\A{The anchor is a hand-specified rule that cannot read news and cannot be improved by outcome-based training; the token policy can do both. The policy finishes within 0.011 of the teacher pooled, surpasses it in 2024 (1.564 vs.\ 1.407), and the modality ablation shows its clearest multimodal advantage in 2025---capabilities unavailable to the optimizer. The paper's claim is feasibility of the LM interface, not that it dominates its own teacher after three test periods.}

\Q{Expectation decoding is a fixed linear readout over bin centers. Isn't ``head-free'' therefore a misnomer?}
\A{The expectation is a parameter-free, post-hoc functional of the emitted token distribution, analogous to reading a probability off a classifier. All trained pathways---token CE, ordinal CE, the rank loss, and the policy gradients---flow through the ordinary tied vocabulary projection, and the discrete tokens remain the sampled, auditable generation trace that the RL stages act on. A regression head, in contrast, adds trained task-specific parameters and a separate loss path and produces no such trace.}

\Q{What is new relative to Pix2Seq-style quantized outputs, LLMTime, TokenCast, and Kronos?}
\A{Those works quantize inputs or outputs for a single task family. The contribution here is one constrained head-free interface covering two different numerical contracts (ordered forecasts and a budgeted long-only action), trained by SFT from a strictly causal financial teacher and then improved by token-level policy optimization against realized portfolio Sharpe. The novelty claim is intentionally kept narrow to this combination; the paper makes no absolute-first claim.}

\Q{The most informative missing ablation is the same backbone with a regression or policy head. Why is it absent?}
\A{We agree it is the most valuable additional experiment and identify it as such. It was deferred because a matched head-based arm requires its own design decisions (head architecture, loss, and a continuous-action RL algorithm), each of which is a confound; the comparisons we do report hold the backbone, prompt, data, grammar, and decoder fixed so that only the training stage or the input modality varies. The present claim is therefore feasibility of the head-free contract, evaluated against calibrated numerical references, not superiority over head-based designs.}

\section{Baselines, Costs, and Interpretation}

\Q{Why is equal weight the only naive baseline? Where are minimum variance, risk parity, and momentum?}
\A{Equal weight is the canonical hard-to-beat 1/N reference, and the causal mean--variance target already provides an optimized covariance-based comparator computed on exactly the information the model sees. Minimum variance and risk parity are close relatives of the same 20-day-window family and are inexpensive to add as further reference rows; we do not expect them to change the SFT-to-policy conclusion, which is internal to the learned models, but agree they would usefully bracket the anchor row.}

\Q{Equal weight earns more raw return in 2024--2025. Is the model just a low-volatility tilt?}
\A{The optimized and rewarded criterion is risk-adjusted return, and equal weight never wins on Sharpe in any period (0.886/1.367/1.506 vs.\ the policy's 1.404/1.564/1.653). Its extra 2025 return comes with 10.03\% volatility and a $-7.99\%$ drawdown against the policy's 6.87\% and $-6.04\%$. Nor is the policy purely defensive: in 2024 and 2025 it raises return over SFT (7.35\%$\to$7.88\%, 10.46\%$\to$11.76\%) while also lowering volatility. A volatility-targeted equal-weight row is a reasonable additional reference we can include.}

\Q{Is the FinTexTS comparison meaningful given the different protocols? Why show it at all?}
\A{The published row is included only to situate the exploratory task, and the paper labels it contextual for precisely this reason. The controlled claim is the within-protocol SFT$\to$policy comparison (73.52\%$\to$73.72\% return, 2.68$\to$2.69 Sharpe) under an identical split, decoder, cost model, and backtest. The divergence between the 19.74-point return gap and the 0.02 Sharpe gap is itself presented as evidence \emph{against} reading the table as cross-system dominance.}

\Q{Are the allocation gains robust to transaction costs beyond the stylized 5 bp?}
\A{Yes, to any plausible linear cost. The policy's pooled net gain (0.100) is nearly identical to its gross gain (0.101) because turnover rises only from 1.481\% to 1.537\% per day. Setting the pooled annual gross-return advantage ($\approx$0.49 pp) against the extra annualized one-way turnover ($0.056\%\times252\approx14.1$ pp) gives a break-even linear cost of roughly 340 bp per unit of one-way turnover---about two orders of magnitude above realistic costs for these five highly liquid ETFs. Impact and slippage models are listed as future work.}

\Q{The policy beats its teacher only in 2024. Is ``regime dependence'' a euphemism for cherry-picking?}
\A{No aggregation was chosen post hoc: the headline pooled statistic is one under which the policy \emph{trails} its own teacher (1.529 vs.\ 1.540), and all three periods are reported with equal prominence. The 2024 crossover demonstrates the mechanism---an outcome reward can push the policy past pure imitation---while 2023 and 2025 bound its size. The paper draws exactly that bounded conclusion.}

\section{Statistical Reliability}

\Q{One training seed and three yearly runs: are the reported gains statistically significant?}
\A{No formal significance is claimed, and the paper says so (``feasibility rather than universal robustness''). The design does, however, make such tests possible: SFT and policy produce paired daily return series on identical dates, so paired Sharpe-difference tests (Jobson--Korkie with the Memmel correction, or a Ledoit--Wolf bootstrap) apply directly to the pooled 709-day series, and repeated seeds are explicitly listed as future work. In the interim, the consistency of the pattern across return, volatility, Sortino, Calmar, and drawdown---rather than a single fragile statistic---is what the interpretation rests on.}

\Q{The 2023 SFT and policy Sharpe are identical to four decimals. Did the policy stage do anything there?}
\A{The two 2023 portfolios are not identical: annual return (7.177\% vs.\ 7.105\%), volatility (5.027\% vs.\ 4.977\%), turnover (1.316\% vs.\ 1.271\%), and drawdown all differ; the return--risk ratio happens to coincide at 1.4042. Behavior-KL early stopping (threshold 0.03) deliberately bounds each policy update, so in a regime where the sampled rewards give little directional signal the optimized policy remains near its initialization. That is the intended failure mode: refine when the signal supports it, otherwise do no harm.}

\Q{The modality-ablation gaps (0.061 and 0.104 in mean Sharpe) could be within single-run noise.}
\A{Agreed, and the paper states that the table ``supports regime-dependent complementarity rather than a statistically general claim.'' What the ablation does establish is structural: with an identical pipeline and only the modality varied, time-series-only nearly ties the multimodal arm in 2023--2024 (gaps of 0.004 and 0.001) while the multimodal arm leads by 0.190 and 0.180 in 2025. The claim is the location of the news contribution, not its universal magnitude.}

\section{Data Integrity and Leakage}

\Q{Llama 3.2's disclosed December 2023 pretraining cutoff overlaps the 2023 allocation test and the entire stock experiment.}
\A{For allocation, the 2024 and 2025 tests postdate the cutoff, and the pooled SFT$\to$policy gain comes entirely from those two periods (2023 contributes zero at three decimals); the paper explicitly flags 2023 for cautious interpretation. For the stock experiment, the 2023 overlap is one reason it is labeled exploratory: the controlled quantity there is the SFT-vs.-policy delta, in which both arms share identical contamination exposure, fine-tuning bins and volatility statistics are computed from pre-2022 data only, and absolute levels are not claimed as contamination-free. Post-cutoff stock evaluation is identified as future work.}

\Q{Daily news is summarized by DeepSeek-V4, a model trained after all test periods. Can the summarizer inject hindsight?}
\A{The summarizer is applied per date to that date's collected articles, but we agree its parametric knowledge is a residual leakage channel worth bounding. Two facts bound it: the time-series-only arm, which contains no news anywhere, reaches 1.565 in 2024 versus 1.564 for the multimodal arm, so the 2024 gain cannot be a news artifact; and the news channel's measurable contribution is confined to 2025 ($+0.190$ over time-series-only), which is therefore the only result a hindsight-bearing summarizer could plausibly inflate. Re-summarizing with a model whose training predates the test periods is a directly feasible robustness check that we flag for the revision.}

\Q{Summarize all look-ahead controls in one place.}
\A{Prompts contain only observations stamped no later than $t$. The mean--variance anchor is recomputed sequentially inside each experiment, so its path-dependent state never crosses from test into training. Policy rewards use returns from $t{+}1$ to $t{+}21$ only, disjoint from every prompt field, and the final 20 training dates are dropped because their reward windows are not fully contained in the training span. Forecast volatility scaling, bin boundaries, and bin centers are estimated from training data only, and ranking losses compare only stocks sharing a decision date. The base-model cutoff is treated separately, as above.}

\Q{Were the anchor constants (0.05 return weight, 0.05 turnover weight, 0.85 shrinkage, 50-unit grid) or the training hyperparameters tuned on the test periods?}
\A{No. All anchor constants and all training settings are single fixed configurations reused unchanged across the three chronological experiments and every ablation arm; nothing is retuned per period, so no test period can influence any selection. The anchor terms follow textbook practice (annualized risk--return tradeoff, a turnover penalty, shrinkage toward 1/N), and the optimization values follow common LoRA and group-relative policy-optimization practice.}

\section{Allocation Design}

\Q{Why GLD, SPY, TLT, UUP, and XLE?}
\A{The five span distinct macro exposures---equities, duration, gold, the dollar, and energy---are among the most liquid ETFs, and share a long common daily history from 2018. This gives the covariance-driven teacher genuine diversification structure and gives the macro-news channel economically distinct hooks. The universe was fixed before evaluation and not selected on test performance.}

\Q{At test time, does the prompt contain the model's own previous allocation or the anchor's previous state? If the latter, the optimizer stays in the loop at deployment.}
\A{Both training and evaluation condition on the previous anchor state $\boldsymbol w_{t-1}^{\anchor}$, recomputed sequentially within each experiment. This state is computable at $t$ from the visible 20-day window alone---a deterministic five-asset program with no future data and no LM in the loop---so the evaluated system is ``LM plus a cheap causal state variable,'' not ``LM plus an oracle.'' The design also gives every arm an identical, well-defined state, which is what makes the SFT/policy and modality comparisons controlled. The fully closed-loop variant, in which the model consumes its own previous output, is an acknowledged next step.}

\Q{Why is the anchor-consistency penalty computed on the raw token action $\boldsymbol z$ rather than the normalized weights, and why coefficient 8? Doesn't this pin the policy to the anchor?}
\A{Penalizing raw grid units teaches budget adherence jointly with composition: deviations in $\boldsymbol z$ capture both misallocation and total-budget errors, which normalization would forgive. The coefficient keeps sampled actions in a trust region around a known-safe teacher so that the very noisy 21-day Sharpe cannot drag the policy anywhere it likes. The 2024 result shows the penalty does not pin the policy: it reaches 1.564 while its anchor achieves 1.407. A sensitivity sweep over the coefficient is future work within the stated single-seed scope.}

\Q{A single 21-day realized Sharpe is extremely noisy. How does learning survive, and why a 21-day reward for a daily-rebalanced portfolio?}
\A{Noise is handled by construction: the group-relative advantage over $G{=}8$ samples of the same prompt removes prompt-level reward level, the standard-deviation floor of 0.65 prevents advantage blow-up in near-tied groups, effectively constant-reward groups are zeroed, the anchor penalty densifies the signal, and three clipped passes with ratio bounds $[0.80,1.28]$ plus behavior-KL early stopping bound each update. Twenty-one trading days is roughly the shortest window with a stable within-window Sharpe---a one-day ``Sharpe'' is undefined---and evaluating each daily action by its month-ahead consequence is a smoothness prior consistent with the anchor's own turnover penalty. The outcome (a 0.10 pooled Sharpe gain at a 5.6 bp/day turnover increase) is consistent with refinement rather than noise-chasing.}

\Q{Consecutive training dates share overlapping 21-day reward windows, correlating rewards.}
\A{Advantages are computed within a group at a single prompt, so the baseline is per-date and the overlap introduces gradient correlation, not bias. One policy epoch with early stopping limits any accumulation of correlated updates. Non-overlapping or horizon-matched reward windows are a straightforward variant for future comparison.}

\Q{Why a 50-unit grid, a 0.5 per-asset cap, and a 50-unit minimum?}
\A{The 21-value vocabulary over $\{0,50,\ldots,1000\}$ balances resolution against sample efficiency: 2.5\% granularity is below realistic implementation noise at daily rebalancing while keeping the ordinal target concentrated on few tokens. The 0.5 cap and the quantizer's per-asset minimum encode standard long-only diversification constraints and ensure the teacher and the policy share one feasible set. All are fixed design constants shared across every experiment.}

\Q{How are invalid or degenerate outputs handled, and how were the $-5$ penalty and 0.65 floor chosen?}
\A{Sampling constrains the answer scaffold so trajectories are grammar-legal; the $-5$ reward is a guard for residual parse failures and sits far below any attainable Sharpe-based reward, so an invalid sample always receives its group's most negative advantage. At deterministic inference the expectation over the fixed value-slot vocabulary is always numerically defined, and the equal-weight fallback is a conservative guard for the degenerate all-near-zero row. Both constants were fixed a priori and shared across all arms; the floor exists because the strong anchor penalty makes near-tied reward groups common.}

\section{Policy Optimization}

\Q{Why is the reference-model KL coefficient zero when GRPO variants usually keep one?}
\A{Following the DAPO observation that reference-KL penalties fight the reward signal in short-horizon structured tasks, the trust region is provided instead by asymmetric clipping, the three-pass limit, and behavior-KL early stopping at 0.03. In addition, the anchor penalty already regularizes the policy in \emph{action} space---a task-meaningful constraint that a distributional KL to a frozen reference only approximates. The paper names the method DAPO-augmented GRPO and tabulates exactly which DAPO mechanisms are and are not present, so no full-DAPO claim is made.}

\Q{Training samples at temperature 1 but evaluation uses deterministic expectation decoding. Is this a train--test mismatch?}
\A{It is the standard on-policy division of labor: sampling provides exploration, and the deployed action is the mean of the learned distribution. Rewards are computed on sampled normalized actions, so sharpening the distribution around high-reward actions moves the deterministic expectation accordingly. Deterministic decoding also removes sampling variance from the evaluation itself, which matters in a single-seed study.}

\section{Forecasting Design}

\Q{Why 41 bins and ordinal width 0.8, and how sensitive are results to these values?}
\A{Forty-one train-only quantile bins place roughly 2.4\% of training mass in each bin, concentrating resolution where standardized returns actually lie, and conditional-mean centers make expectation decoding calibrated on the training distribution by construction. Width 0.8 spreads ordinal credit over adjacent bins so near-misses are graded rather than punished as full errors. Both were fixed a priori; sensitivity sweeps are future work within the stated scope.}

\Q{The backtest is a top-20 strategy, yet the top-20 reward component gets weight only 0.05. Why the mismatch?}
\A{The backtest depends on the cross-sectional ordering, which the 0.25 rank component targets directly and densely, whereas realized top-20 basket return is a sparse, high-variance signal that invites selection-noise chasing if weighted heavily. The step and path components (0.65 combined) preserve calibration of the underlying scores that generate the ranking. The small top-20 term keeps the deployed criterion visible to the policy without letting it dominate.}

\Q{Why is the forecasting policy gain so small (73.52\%$\to$73.72\%, 2.68$\to$2.69)?}
\A{The one-epoch stage with a 0.02 KL to cached SFT probabilities is deliberately conservative, and the paper interprets the result accordingly: the supervised token representation already carries most of the predictive signal, and the policy stage supplies incremental task alignment without destabilizing it. The design claim being tested is that the token interface \emph{tolerates} outcome-based RL at all---which it does, improving both reported metrics---not that RL is the main source of performance.}

\Q{How are the numeric tokens implemented?}
\A{All task tokens---41 bin tokens plus horizon and boundary markers for forecasting, and 21 grid values plus five ETF tags for allocation---are added vocabulary items scored by the tied input--output embedding, with no separate head. They are trained jointly with the LoRA adapters during SFT, and in the forecasting policy stage the special embeddings receive their own learning-rate schedule ($5\times10^{-6}\!\to\!5\times10^{-7}$) alongside the policy rate ($10^{-6}\!\to\!10^{-7}$).}

\Q{Why does allocation SFT apply cross-entropy to the full answer, while forecasting trains only the three bucket slots under a fixed scaffold?}
\A{The allocation answer's tags and ordering are part of the contract the model must reproduce, so full-sequence CE teaches the entire grammar; the forecasting scaffold is a constant frame with no content, so it is forced and the losses concentrate on the three informative slots. Both choices avoid spending capacity on constants while ensuring the emitted trace is fully specified and mechanically parseable.}

\Q{Which pairs enter the ranking loss?}
\A{Only pairs of stocks sharing the same decision date, formed per horizon, with coefficient 0.5. Pairs never span dates, so the loss trains the score for exactly its cross-sectional ranking use without introducing temporal comparisons that could leak ordering information across time.}

\section{Practicality and Reproducibility}

\Q{What are the compute and latency costs? Is a daily LLM allocation loop practical?}
\A{Inference is one constrained generation of roughly a dozen answer tokens per day on a 1B-parameter LoRA model in bfloat16---negligible next to any daily data pipeline and feasible on a single GPU. Training per experiment is five SFT epochs over at most about five years of daily prompts plus one policy epoch. The head-free design adds no inference-time machinery beyond a softmax expectation at the value slots.}

\Q{Does the interface extend to new assets or universes without full retraining?}
\A{The bin and grid vocabularies are asset-agnostic because targets are volatility-standardized or budget-normalized; the ETF tags and the anchor are universe-specific, so adding assets means extending the tag set and re-running LoRA SFT, which is cheap. The forecasting model already shares one vocabulary across 100 stocks, which is within-universe evidence of cross-sectional generality; transfer across universes is untested and is listed as a limitation.}

\Q{Why Llama 3.2 1B? Would scale, or a randomly initialized backbone, change the conclusions?}
\A{The 1B model is the smallest widely available causal LM that comfortably ingests the long multimodal prompt, which keeps every ablation affordable and makes the feasibility claim conservative: the interface does not depend on frontier-scale capability. Scale and random-initialization ablations (probing how much the pretrained LM prior contributes) are natural next steps; the fact that the news channel produces the 2025 advantage suggests the text-reading capability---an LM-specific asset---is doing real work.}

\end{document}